%% file: ACG2025.tex
\documentclass[runningheads]{llncs}
\usepackage[T1]{fontenc}
\usepackage{graphicx}
\usepackage{algorithm}
\usepackage{algorithmic}
\usepackage{booktabs}
\usepackage{caption}
\usepackage{adjustbox}
\usepackage{csvsimple}
\usepackage{pgfplots}
\usepackage{pgfplotstable}
\pgfplotsset{compat=1.17}
\usepackage[absolute,overlay]{textpos}
\usepackage{multirow}
\usepackage{makecell}
\usepackage{comment}
\usepackage{svg}
\usepackage{CJKutf8}
\usepackage{amsmath}
\usepackage{subcaption}
\usepackage{amssymb}

\begin{document}
\title{Relevance-Zone Reduction in Game Solving}
%
%
\author{Chi-Huang Lin\inst{1}\orcidID{0009-0000-5078-7866} \and
Ting Han Wei\inst{2}\orcidID{0009-0004-6060-1905} \and
Chun-Jui Wang\inst{1}\orcidID{0009-0002-8728-2508} \and
Hung Guei\inst{3}\orcidID{0000-0002-5590-7529} \and
Chung-Chin Shih\inst{3}\orcidID{0000-0003-4261-4871} \and 
Yun-Jui Tsai\inst{1}\orcidID{0009-0007-6703-1687} \and
I-Chen Wu\inst{1,3}\orcidID{0000-0003-2535-0587} \and
Ti-Rong Wu\inst{3}\orcidID{0000-0002-7532-3176}
}
\authorrunning{Lin et al.}
%
\institute{National Yang Ming Chiao Tung University, Hsinchu, Taiwan \and
Kochi University of Technology, Kami City, Japan
\\
\and
Academia Sinica, Taipei, Taiwan
\\
\email{tirongwu@iis.sinica.edu.tw}
}

\maketitle              

\input{main}
\bibliographystyle{splncs04}
\bibliography{reference}

\end{document}

%% file: main.tex
\begin{abstract}
Game solving aims to find the optimal strategies for all players and determine the theoretical outcome of a game.
However, due to the exponential growth of game trees, many games remain unsolved, even though methods like AlphaZero have demonstrated super-human level in game playing.
The Relevance-Zone (RZ) is a local strategy reuse technique that restricts the search to only the regions relevant to the outcome, significantly reducing the search space.
However, RZs are not unique. Different solutions may result in RZs of varying sizes.
Smaller RZs are generally more favorable, as they increase the chance of reuse and improve pruning efficiency.
To this end, we propose an iterative RZ reduction method that repeatedly solves the same position while gradually restricting the region involved, guiding the solver toward smaller RZs.
We design three constraint generation strategies and integrate an RZ Pattern Table to fully leverage past solutions.
In experiments on $7 \times 7$ Killall-Go, our method reduces the average RZ size to 85.95\% of the original. Furthermore, the reduced RZs can be permanently stored as reusable knowledge for future solving tasks, especially for larger board sizes or different openings.
    \keywords{Game solving \and Relevance-Zone \and Pattern Table \and Killall-Go.}
\end{abstract}

\section{Introduction}
Game solving is a similar, yet different task than game playing \cite{vandenherik_games_2002}.
To illustrate, super-human level play for various games can be achieved without requiring human knowledge using the AlphaZero \cite{silver_general_2018} and MuZero \cite{schrittwieser_mastering_2020} algorithms.
However, these programs can still make mistakes by playing sub-optimal moves, whether through inaccurate evaluation \cite{haque_road_2022} or through adversarial attacks and exploits \cite{wang_adversarial_2023,lan_are_2022}.
On the other hand, game solving focuses on finding the outcome of a game given optimal play for all players, which is known as the \textit{game-theoretic value}.
If this game-theoretic value is obtained by brute-force search, a \textit{solution tree} that records all player strategies can be constructed, guaranteeing perfect play. 
However, given the exponential growth of game trees, many games cannot be feasibly analyzed without the use of carefully designed heuristics or improvements in search algorithms.

To reduce the size of the solution tree, a key idea is to reuse solutions for identical board states, or \textit{transpositions}.
A table of game-theoretic values for transpositions can be maintained, significantly reducing the search size, facilitating full analysis of relatively complicated games.
A generalization of this idea is reusing strategies for partial pattern matches, called a \textit{Relevance-Zone} (RZ) \cite{shih_novel_2022}, instead of full board matches.
An RZ is a region that contains all points that need to be considered to replay the strategy leading to its respective game-theoretic value.
In other words, moves made outside the RZ can be safely pruned without affecting the result of the search.
RZ pattern tables (RZT) \cite{shih_localpattern_2023} have been effective for games including Hex, Go, and Killall-Go \cite{shih_novel_2022}.

An important detail is that there may be multiple winning strategies for the same position, resulting in different RZ patterns.
More specifically, in the worst case, the RZ covers the entire board and is equivalent to the transposition.
Given the choice between multiple RZ patterns, smaller RZs are preferred for two reasons.
First, more moves can be pruned since there are fewer moves inside the RZ.
Second, more matches can be made with smaller patterns, meaning more positions can be skipped during search.
Therefore, finding a strategy that results in a smaller RZ can enhance overall search performance.

In this paper, we propose an iterative RZ reduction algorithm.
By repeatedly re-solving the same position and gradually guiding the solution away from playing in certain regions, our method generates smaller RZs over time.
We design several constraint generation strategies and integrate an RZT to reuse past solutions to accelerate the solving process.
Experiments on 10,000 opening positions in $7 \times 7$ Killall-Go show that our method reduces the RZ to $85.95\%$ of its original size on average, leading to more efficient strategy reuse.
Moreover, our RZ reduction efforts can be stored permanently to help future game solving efforts.
This means that the offline time spent generating the RZT can be repaid multiple times when attempting to solve Go or Killall-Go for bigger board sizes.

\section{Background}
\subsection{Relevance-Zones and Pattern Tables}
In board games, players typically engage in battle within localized regions of the board.
Intuitively, moves far away from this active region are unlikely to affect the outcome of the game.
Relevance-Zones (RZs) \cite{shih_novel_2022,thomsen_lambdasearch_2001,wu_relevancezoneoriented_2010} are an idea that formally defines this critical region.
RZs are masks on the game board such that any number of moves made outside of this region do not affect the game-theoretic value of the position.
RZs can be generated automatically during search.
When a terminal position is examined, the stone configuration within the region that contributed to the result is called the RZ pattern (key), and its corresponding outcome (value) is saved into the \textit{RZ pattern table} (RZT) \cite{shih_localpattern_2023}.
RZT entries can also be propagated upward to the terminal position's ancestors, with careful consideration to the rules of the game.
This propagation is subject to the parent node's location in an AND-OR tree.
For an AND node, once all relevant child nodes are solved, their RZ patterns are merged in a union and updated.
For an OR node, the RZ of any solved child can be directly updated, often expanded according to the game's rules.

During the search, for every solved position, a newly generated RZ pattern is stored.
The table is queried every time a node is expanded.
If a matching RZT entry is found, the node is immediately marked as solved and does not require further expansion.

\subsection{$7 \times 7$ Killall-Go Solver}
Killall-Go is a variant of Go, where White only needs to secure a single safe region to win, while Black must kill all white stones to win.
In $7 \times 7$ Killall-Go, Black is allowed to place two stones for their first move as a compensating advantage.
Since White's objective is to form living groups, the entire game can be viewed as a life-and-death problem.
Many Go experts believe that $7 \times 7$ Killall-Go is a win for White \cite{wu_game_2024}.
Therefore, in this study, we treat White as the OR-player and attempt to prove its guaranteed win.

AlphaZero \cite{silver_general_2018} demonstrated the possibility of super-human level play without human knowledge by maximizing the win rate or board evaluation during search.
To solve $7 \times 7$ Killall-Go, Wu et al. \cite{wu_alphazerobased_2021} proposed using AlphaZero-style training to create heuristics for game solving by modifying the training target of the value network to predict the size of the solution instead, calling the modified network the \textit{Proof Cost Network} (PCN).
Experiments showed that the use of PCNs required significantly fewer nodes to complete proofs for $15 \times 15$ Gomoku and $9 \times 9$ Killall-Go.

The state-of-the-art for solving $7 \times 7$ Killall-Go integrates RZTs and the PCN \cite{wu_game_2024}. 
Additionally, it adopts a manager-worker distributed computing paradigm while also leveraging online fine-tuning during solving.
Namely, the PCN is continuously adjusted based on the current subtree being examined, allowing the neural network to produce more accurate proof cost predictions in real time.
The resulting solver was able to solve 16 selected three-move opening positions, searching only $21.69\%$ nodes than the baseline solver without online fine-tuning.
The full solution trees for these positions were also published.

In this paper, we use the worker component of this online fine-tuning distributed solver as our solver, and select a subset of their publicly released solution trees as benchmark openings for our experiments.

\section{Relevance-Zone Reduction}
\subsection{Algorithm}
RZs were designed for local pattern matching: when two board positions share the same pattern inside their RZ, the solution computed for one position can be reused for the other.
The smaller the RZ, the more likely a different position might share that pattern, increasing the chance of reuse, speeding up the solving process, and reducing the size of the overall solution tree.

A single board position may have multiple valid solutions, i.e. its corresponding solution tree and RZ are not unique.
However, from a practical point of view, the solver is satisfied with obtaining any solution, even if smaller solution trees may be possible.
In fact, previous studies have demonstrated that computing a minimum solution graph is computationally intractable \cite{plaat_nearly_2014}.
Indeed, in practice, solution trees are constructed in a bottom-up manner, with no simple way of guaranteeing the resulting graph or the RZ patterns themselves are minimal, short of enumerating all possible sub-graphs.
We therefore propose an iterative RZ reduction approach \textit{RZR}, attempting to solve the same position multiple times, progressively reducing the RZ size in each iteration.

\begin{figure}[h]
\centering
\includegraphics[width=1.0\columnwidth]{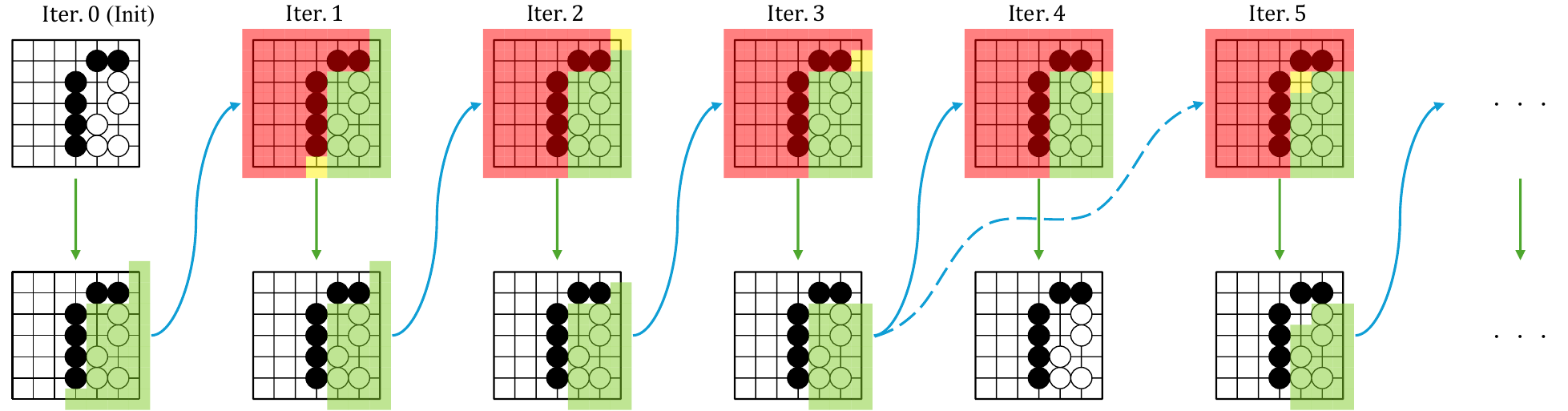}
\caption{Workflow of iterative RZ reduction (RZR).
Top-row: constraint for each round—RZ from the last solved round (green), locked outer-region (red), newly banned point (yellow, overlap with RZ).
Bottom-row: RZ returned by the solver under that constraint (unshaded if unsolved).
}
\label{fig:architecture}
\end{figure}

To do this, we introduce the constraint of prohibited regions on the board.
In each iteration, the solver is required to return a solution whose RZ does not overlap with this prohibited region.
By blocking parts of the board, the constraint rules out larger RZs and ensures that each successful RZ is smaller than its predecessors.

The complete procedure has two phases: \textbf{initialization} and \textbf{iterative re-solving}.
In the initialization phase, the solver first tackles the position without any constraints.
If it succeeds, the resulting solution tree becomes the baseline for later rounds; if it fails, the position is declared unsolved.
Unsolved positions will be discarded and will not proceed to the iteration phase.
During the iteration phase, each new round builds a constraint from the root-level RZ of the most recent successful run, then re-solves the position under that constraint.
The procedure is as follows: 

\begin{enumerate}
    \item \textbf{Outer-region lock:} (red region in Fig. \ref{fig:architecture}) \\
        Every point outside the previous RZ is added to the constraint to prevent subsequent attempts from expanding outward.
    \item \textbf{Inner-point removal:} (yellow region in Fig. \ref{fig:architecture}) \\
        Exactly one previously unused point inside the RZ is also added to the constraint.
        This forces the solver to find a solution that does not rely on that point, further shrinking the region.
    \item \textbf{Re-solve:} (green arrow in Fig. \ref{fig:architecture}) \\
        The position and its constraint are given to the solver.
        If no legal solution is found that avoids the constraint within a fixed simulation budget, the constraint is marked as failed.
    \item \textbf{Caching and reuse (RZT):} \\
        All RZ patterns discovered in the round are stored in the RZT.
        Later searches can match these patterns directly to avoid search redundancies.
    \item \textbf{Skipping hopeless points}: \\
        When a constraint fails, the inner point chosen above is recorded so that future rounds will not waste time on it again.
\end{enumerate}

To prevent wasting excessive simulations on overly restrictive constraints, we impose a fixed simulation budget for every re-solve attempt.
If the solver exhausts this budget without finding a legal solution that respects the constraint, the constraint is deemed to have failed.
The reduction cycle then repeats until no smaller valid RZs can be obtained in $K$ consecutive iterations.
The algorithm then terminates, thereby avoiding further resource usage on a position that is unlikely to shrink.
The smallest RZ pattern discovered during the entire process is the output.

\subsection{Constraint Generation}
A constraint is a specified region on the board that restricts where the RZ is allowed to form.
The solver receives this constraint at the start of each iteration.
While re-solving, any leaf node whose RZ overlaps with the constraint is immediately considered unsolved.
This forces the solver to search for an alternative solution tree that avoids such nodes.
If the constraint is too strict, it may become impossible to find a valid solution tree.
Therefore, choosing which point within the RZ to add to the constraint is a central challenge in the reduction algorithm.
We propose three strategies for constraint point selection:
\begin{enumerate}
    \item \textbf{Random:} Randomly sample a point uniformly from the current RZ.
    \item \textbf{Erosion:} Considers only empty points inside the RZ, and ranks them by how many of their neighboring points lie outside the RZ.
    Points with more external neighbors are prioritized.
    \item \textbf{Heatmap:} Aggregates the RZs from all nodes in the current solution tree to create a heatmap, where each point reflects how many RZs cover it.
    The point with the lowest heat (i.e., least covered) is selected next to avoid simultaneously invalidating too many RZs in a single iteration.
\end{enumerate}

\subsection{Modifications to the Relevance-Zone Pattern Table (RZT)}
The RZT significantly reduces unnecessary search overhead, allowing the solver to focus on yet-uncovered, critical subtrees.
In the implementation by Shih et al. \cite{shih_localpattern_2023}, the goal was to find any valid solution, so any matching pattern can be returned for each RZT query.
However, since solutions are not unique, multiple RZs with different shapes and sizes might correspond with the same position.
Thus, when performing RZR, RZT queries return all matching entries.
We filter out all patterns that violate the current constraint, then select the smallest valid RZ to maximize future reuse and speed up solving.

Furthermore, since all known RZs are stored across different iterations, the overall solution tree tends to progressively shrink.
With each new round, the RZT becomes more populated with reusable local patterns, covering more board positions, and improving the likelihood of finding constraint-compliant alternatives.
In contrast, disabling the RZT forces each iteration to rebuild the entire tree from scratch, leading to redundant computation and dramatically reduced efficiency.

\section{Experiments}
\subsection{Setup}
\begin{figure}[h]
    \captionsetup[subfigure]{labelformat=empty}
    \centering
    \subfloat[JA]{
        \includegraphics[width=0.17\columnwidth]{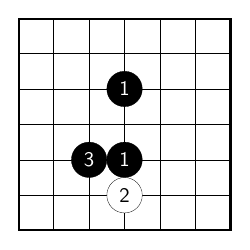}
        \label{fig:JA}
    }
    \subfloat[KA]{
        \includegraphics[width=0.17\columnwidth]{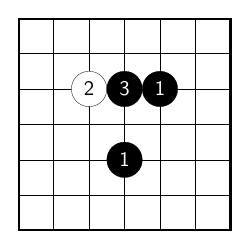}
        \label{fig:KA}
    }
    \subfloat[DA]{
        \includegraphics[width=0.17\columnwidth]{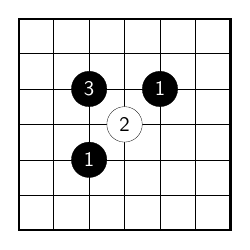}
        \label{fig:DA}
    }
    \subfloat[SA]{
        \includegraphics[width=0.17\columnwidth]{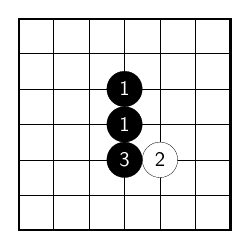}
        \label{fig:SA}
    }
    \caption{Four openings from \cite{wu_game_2024}.}
    \label{openings}
\end{figure}

In our experiments, we use the worker component from the Online Fine-Tuning Distributed Solver \cite{wu_game_2024} as our baseline solver.
The proposed RZR is built on top of this solver by adding constraints to limit the region where RZs can be formed.
Since each position is re-solved multiple times, we avoid redundant evaluations by caching PCN outputs in a hash table.
From the 16 published ONLINE‑CP solution trees \cite{wu_game_2024}, we select four opening positions: JA, KA, DA, and SA, as shown in Fig. \ref{openings}.
For each opening, we randomly sample 2,500 leaf nodes as benchmark problems.
Among them, KA and SA are relatively easier (with smaller solution trees), while JA and DA are more challenging.

During the initialization phase, we set the solver’s simulation count limit to 100,000, matching the configuration in the original Online Fine-Tuning solver.
The size of the RZ obtained in this phase is used as the baseline for comparison.
In the iterative re-solving phase, we reduce the simulation count limit to 20,000.
We found that most problems can still be solved under this setting, which helps avoid wasting resources on overly strict constraints that lead to unsolvable cases.
Additionally, if no successful solution is found for five consecutive iterations (i.e., $K=5$), the process is terminated.
All experiments are performed on a machine with a 1080 Ti GPU.

\subsection{Improvement}
In this section, we analyze how different constraint generation strategies and the use of the RZT affect the final RZ size under a fixed setting of $K = 5$.
As shown in Table 1, the combination of Heatmap and RZT achieves the best performance, reducing the RZ size to an average of $85.95\%$ of the original.
It also achieves the best results across all four openings.
In contrast, the Random strategy performs the worst, reducing the RZ size only to $92.33\%$ on average.

In addition, using the RZT leads to better results across all three constraint generation methods.
By allowing reuse of previously solved RZs that satisfy the current constraint, the same node may have multiple RZ options, from which we can choose the smallest.
Furthermore, nodes that are skipped do not consume computing resources, increasing the chance of finding a valid solution.

\input{table/table1}

\begin{figure}[h]
\centering
\includegraphics[width=0.6\columnwidth]{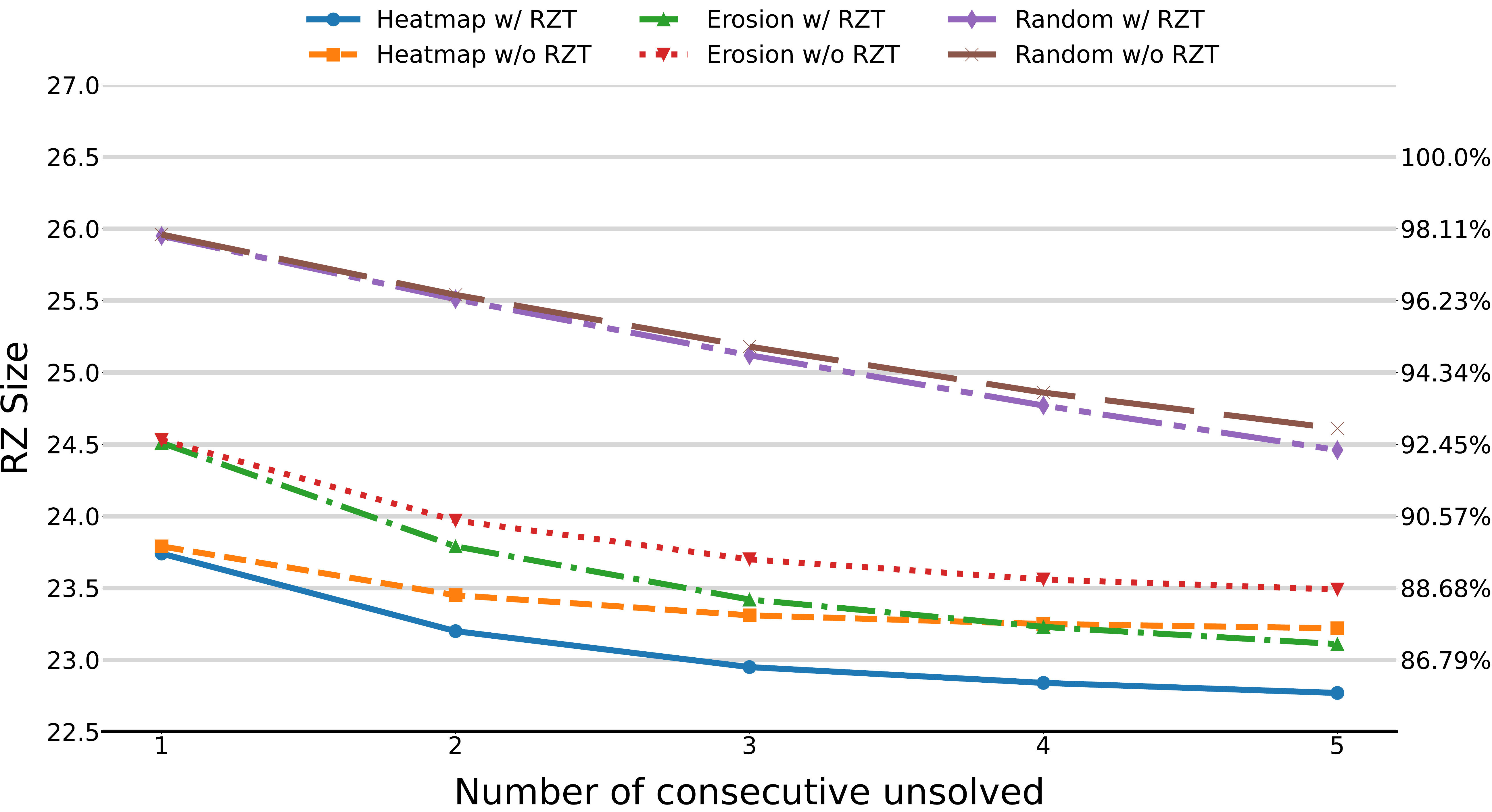}
\caption{Early Stopping.}
\label{fig:early-stopping}
\end{figure}
We also analyze the effect of different early stopping thresholds $K$.
Fig. \ref{fig:early-stopping} shows the corresponding RZ reduction rates for varying values of $K$.
Even when we stop RZR immediately after the first failure ($K=1$), the average RZ size already drops noticeably.
By $K = 5$, the heatmap strategy achieves an average reduction rate close to $86\%$, requiring about 8.2 iterations on average to terminate.
Beyond this point, increasing $K$ by 1 incurs an additional computational cost of approximately $12\%$ compared to the previous setting, while the RZ size decreases by only $0.2\%$.
This indicates diminishing returns and suggests that larger $K$ values offer limited practical benefit.

\input{table/table2}
To further understand the impact of RZ size, we group the data based on the baseline RZ into four ranges.
As shown in Table 2, positions with larger original RZs tend to achieve higher reduction ratios.
In contrast, positions with an initial RZ size below 20 are reduced by only about one point on average.
Additionally, enabling or disabling the RZT has little impact when the RZ is already small.
This is because the main benefit of RZTs comes from reusing solutions across different iterations.
When the reduction ratio is low, most iterations tend to fail early, resulting in fewer reusable patterns stored in the RZT.

\subsection{Case Studies}
\input{figure/irreducible}
In this section, we analyze three representative positions to better understand the effect of RZR. 
Fig. \ref{fig:irreducible} presents two examples of irreducible RZs. These include a DA position where the RZ has been successfully reduced, and an SA position that was irreducible from the start.
In the DA example, the yellow region was successfully excluded by RZR.
The points labeled 13, A, and B are necessary for White to form a two-eye live group and thus must remain within the RZ.
Points A and B form a \textit{miai} pair; no matter where Black plays, White can respond with the other to capture the Black stone and live without needing points C or D.
The key difference in RZ size stems from White's move choice between the positions marked $\times$ and $\triangle$.
Both moves lead to a live group, so the PCN cannot easily distinguish between them.
However, RZR imposes constraints that guide the solver to favor the path resulting in a smaller RZ.
For the irreducible SA position, points A and B form a miai pair. To maintain this miai, both A and B must be empty at the same time, so they must be included within the RZ.

\input{figure/variations}
Fig. \ref{fig:variations} shows the example from the DA opening that achieved the largest RZ reduction.
The four cases shown represent key move differences between the baseline and RZR solution trees, specifically in White’s move choices.
The second and third columns display how the game unfolds after each move.
We observe that in the baseline, the different move choices result in four RZs located in completely different parts of the board.
As a result, when their union is taken, the RZ covers the entire board.
In contrast, the RZR-selected moves lead to RZs that are all concentrated in the lower-left corner.
Their union is also localized in that region, significantly reducing the final RZ size.

\section{Conclusion}
This paper proposes an iterative Relevance-Zone (RZ) reduction method that guides the solver to produce smaller RZs by repeatedly solving the same position under region-based constraints.
We introduce three constraint selection strategies and integrate an RZ Pattern Table (RZT) to accelerate search through reuse.
Experiments show that our method reduces the average RZ size to 85.95\% of the original.
Furthermore, the reduced RZs can be permanently stored as reusable knowledge for future solving tasks.
These RZs can be applied to the full $7 \times 7$ Killall-Go solving, as well as to standard Go rules or different board sizes.

\begin{credits}
\subsubsection{\ackname}

This research is partially supported by the National Science and Technology Council (NSTC) of the Republic of China (Taiwan) under Grant Numbers 113-2221-E-001-009-MY3, 113-2634-F-A49-004, 114-2221-E-A49-005, and 114-2221-E-A49-006.

\end{credits}

%% file: table/table1.tex
\begin{table}[ht]
\caption{RZ size in number of points, reduction rate, grouped by opening.}
    \centering
    \setlength{\tabcolsep}{5pt}
    \begin{adjustbox}{width=\columnwidth}
    \begin{tabular}{crr r rr r rr r rr}
    \toprule
    \multirow{2}{*}{Opening} & \multirow{2}{*}{\# Problems} & \multicolumn{1}{c}{Baseline} && \multicolumn{2}{c}{Heatmap} && \multicolumn{2}{c}{Erosion}  && \multicolumn{2}{c}{Random}\\
    \cline{5-6}
    \cline{8-9}
    \cline{11-12}
    & & \multicolumn{1}{c}{(Avg. Size)} && \multicolumn{1}{c}{w/ RZT} & \multicolumn{1}{c}{w/o RZT} && \multicolumn{1}{c}{w/ RZT} & \multicolumn{1}{c}{w/o RZT} && \multicolumn{1}{c}{w/ RZT} &  \multicolumn{1}{c}{w/o RZT} \\
    \midrule
       JA   &    2,500    &  29.29   &&    \textbf{25.26 (86.23\%)}  &  26.01 (88.79\%)  &&  25.71 (87.80\%)  &  26.25 (89.61\%)  &&  27.16 (92.75\%) &  27.43 (93.67\%) \\
       KA   &    2,500    &  25.03   &&    \textbf{21.32 (85.18\%)}  &  21.77 (86.99\%)  &&  21.66 (86.55\%)  &  22.07 (88.19\%)  &&  22.95 (91.70\%) &  23.10 (92.30\%) \\
       DA   &    2,500    &  28.20   &&    \textbf{24.37 (86.40\%)}  &  24.84 (88.08\%)  &&  24.76 (87.78\%)  &  25.21 (89.38\%)  &&  26.19 (92.85\%) &  26.29 (93.23\%) \\
       SA   &    2,500    &  23.44   &&    \textbf{20.14 (85.91\%)}  &  20.28 (86.51\%)  &&  20.32 (86.67\%)  &  20.45 (87.25\%)  &&  21.53 (91.86\%) &  21.60 (92.14\%) \\
    \bottomrule
       All  &    10,000   &  26.49   &&    \textbf{22.77 (85.95\%)}  &  23.22 (87.67\%)  &&  23.11 (87.25\%)  &  23.49 (88.69\%)  &&  24.46 (92.33\%) &  24.61 (92.89\%) \\
    \bottomrule

    \end{tabular}
    \end{adjustbox}
    \label{tab:table-information-1}
\end{table}

%% file: table/table2.tex
\begin{table}[ht]
\caption{RZ size in number of points, reduction rate, grouped by RZ size.}
    \centering
    \setlength{\tabcolsep}{5pt}
    \begin{adjustbox}{width=\columnwidth}
    \begin{tabular}{crr r rr r rr r rr}
    \toprule
    \multirow{2}{*}{RZ Size} & \multirow{2}{*}{\# Problems} & \multicolumn{1}{c}{Baseline}  && \multicolumn{2}{c}{Heatmap} && \multicolumn{2}{c}{Erosion} && \multicolumn{2}{c}{Random} \\
    \cline{5-6}
    \cline{8-9}
    \cline{11-12}
          &             & \multicolumn{1}{c}{(Avg. Size)} && \multicolumn{1}{c}{w/ RZT} & \multicolumn{1}{c}{w/o RZT} && \multicolumn{1}{c}{w/ RZT} & \multicolumn{1}{c}{w/o RZT} && \multicolumn{1}{c}{w/ RZT} &  \multicolumn{1}{c}{w/o RZT} \\
    \midrule
         $\le$20 & 2,147            (21.47\%) & 17.27 && 16.18 (93.66\%) & \textbf{16.17 (93.63\%)} && 16.22 (93.93\%) & 16.22 (93.93\%) && 16.74 (96.90\%) & 16.70 (96.68\%) \\
         21-30   & 5,190            (51.90\%) & 25.39 && \textbf{22.06 (86.89\%)} & 22.15 (87.22\%) && 22.31 (87.87\%) & 22.39 (88.20\%) && 23.61 (92.98\%) & 23.64 (93.09\%) \\
         31-40   & 2,213            (22.13\%) & 34.45 && \textbf{28.07 (81.49\%)} & 29.22 (84.80\%) && 28.72 (83.36\%) & 29.70 (86.21\%) && 30.82 (89.47\%) & 31.21 (90.58\%) \\
         >40     &   450 \hphantom{0}(4.50\%) & 44.00 && \textbf{36.31 (82.53\%)} & 39.83 (90.52\%) && 37.67 (85.63\%) & 40.36 (91.73\%) && 39.84 (90.55\%) & 41.06 (93.32\%) \\
    \bottomrule
    \end{tabular}
    \end{adjustbox}
    \label{tab:table-information-2}
\end{table}

%% file: figure/irreducible.tex


\begin{figure}[ht]
    \captionsetup[subfigure]{justification=centering}
    \captionsetup[subfigure]{labelformat=empty}
    \centering
    \subfloat[\hspace{0.15cm}DA variant]{
        \includegraphics[width=0.17\columnwidth]{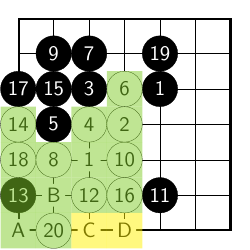}
    }
    \subfloat[\hspace{0.15cm}w/o RZR]{
        \includegraphics[width=0.17\columnwidth]{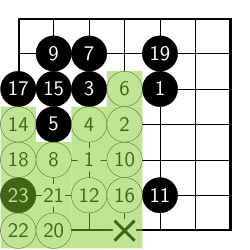}
    }
    \subfloat[\hspace{0.15cm}w/ RZR]{
        \includegraphics[width=0.17\columnwidth]{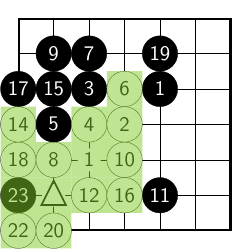}
    }
    \hspace{1.0cm}
    \subfloat[\hspace{0.15cm}SA variant]{
        \includegraphics[width=0.17\columnwidth]{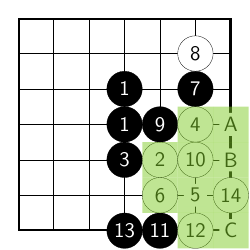}
    }
    \caption{Irreducible positions.}
    \label{fig:irreducible}
\end{figure}


%% file: figure/variations.tex
\begin{figure}[ht]
    \centering
    \begin{tabular}{>{\raggedleft\arraybackslash}m{1.0cm}m{0.17\columnwidth}!{\vrule width 1pt}*{4}{m{0.17\columnwidth}}}
        &
        \multicolumn{1}{c}{} &
        \multicolumn{1}{c}{Case 1} &
        \multicolumn{1}{c}{Case 2} &
        \multicolumn{1}{c}{Case 3} &
        \multicolumn{1}{c}{Case 4} \\
        &
        \includegraphics[width=\linewidth]{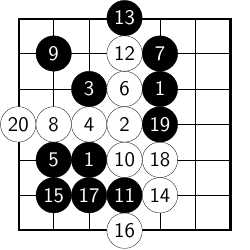} &
        \includegraphics[width=\linewidth]{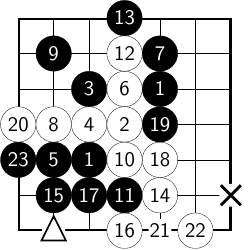} &
        \includegraphics[width=\linewidth]{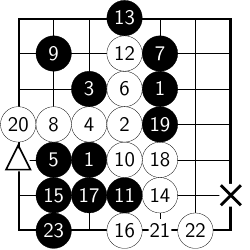} &
        \includegraphics[width=\linewidth]{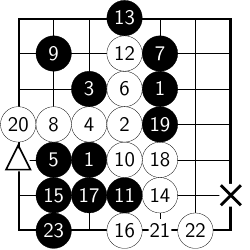} &
        \includegraphics[width=\linewidth]{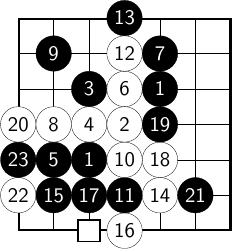} \\

        \makecell{w/o \\ RZR} &
        \includegraphics[width=\linewidth]{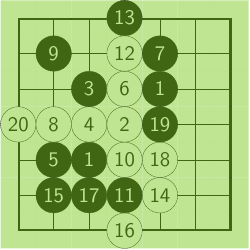} &
        \includegraphics[width=\linewidth]{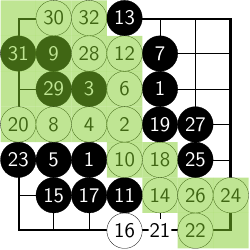} &
        \includegraphics[width=\linewidth]{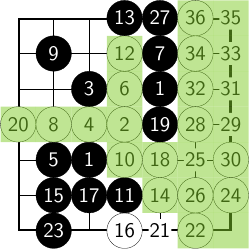} &
        \includegraphics[width=\linewidth]{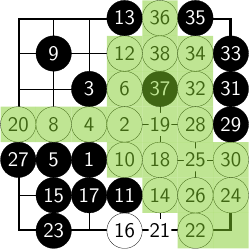} &
        \includegraphics[width=\linewidth]{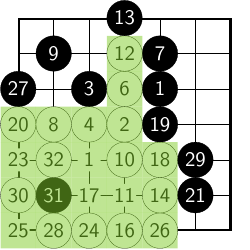} \\

        \makecell{w/ \\ RZR} &
        \includegraphics[width=\linewidth]{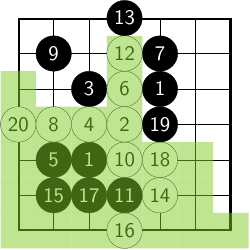} &
        \includegraphics[width=\linewidth]{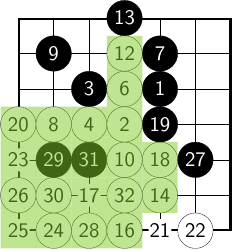} &
        \includegraphics[width=\linewidth]{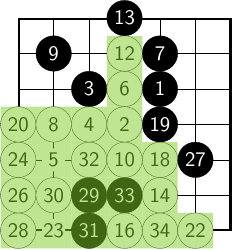} &
        \includegraphics[width=\linewidth]{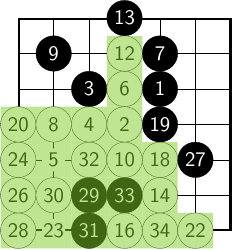} &
        \includegraphics[width=\linewidth]{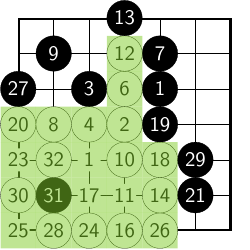}
    \end{tabular}
    \caption{Sample problem from the DA opening.  
Initial position (upper‑left); root RZs w/o and w/ RZR (middle‑left, bottom‑left).  
Columns 1–4: representative branches—$\times$ baseline white move, $\triangle$ RZR‑guided move (top row).  
Shaded cells mark the minimal leaf‑level RZs; their union produces the RZs in Column 1. Moves 25 and 31 are played at the same point in the third row, third column. For Case 4, both the baseline and RZR play at the same position ($\Box$).}
    \label{fig:variations}
\end{figure}

%% file: ACG2025.bbl
\begin{thebibliography}{10}
\providecommand{\url}[1]{\texttt{#1}}
\providecommand{\urlprefix}{URL }
\providecommand{\doi}[1]{https://doi.org/#1}

\bibitem{haque_road_2022}
Haque, R., Wei, T.H., M{\"u}ller, M.: On the~{{Road}} to~{{Perfection}}? {{Evaluating Leela Chess Zero Against Endgame Tablebases}}. In: Advances in {{Computer Games}}. pp. 142--152. Lecture {{Notes}} in {{Computer Science}}, Springer International Publishing, Cham (2022)

\bibitem{lan_are_2022}
Lan, L.C., Zhang, H., Wu, T.R., Tsai, M.Y., Wu, I.C., Hsieh, C.J.: Are {{AlphaZero-like Agents Robust}} to {{Adversarial Perturbations}}? Advances in Neural Information Processing Systems  \textbf{35},  11229--11240 (Dec 2022)

\bibitem{plaat_nearly_2014}
Plaat, A., Schaeffer, J., Pijls, W., de~Bruin, A.: Nearly {{Optimal Minimax Tree Search}}? (Apr 2014)

\bibitem{schrittwieser_mastering_2020}
Schrittwieser, J., Antonoglou, I., Hubert, T., Simonyan, K., Sifre, L., Schmitt, S., Guez, A., Lockhart, E., Hassabis, D., Graepel, T., Lillicrap, T., Silver, D.: Mastering {{Atari}}, {{Go}}, chess and shogi by planning with a learned model. Nature  \textbf{588}(7839),  604--609 (Dec 2020)

\bibitem{shih_localpattern_2023}
Shih, C.C., Wei, T.H., Wu, T.R., Wu, I.C.: A {{Local-Pattern Related Look-Up Table}}. IEEE Transactions on Games pp. 1--10 (2023)

\bibitem{shih_novel_2022}
Shih, C.C., Wu, T.R., Wei, T.H., Wu, I.C.: A {{Novel Approach}} to {{Solving Goal-Achieving Problems}} for {{Board Games}}. In: Proceedings of the {{AAAI Conference}} on {{Artificial Intelligence}}. vol.~36, pp. 10362--10369 (Jun 2022)

\bibitem{silver_general_2018}
Silver, D., Hubert, T., Schrittwieser, J., Antonoglou, I., Lai, M., Guez, A., Lanctot, M., Sifre, L., Kumaran, D., Graepel, T., Lillicrap, T., Simonyan, K., Hassabis, D.: A general reinforcement learning algorithm that masters chess, shogi, and {{Go}} through self-play. Science  \textbf{362}(6419),  1140--1144 (Dec 2018)

\bibitem{thomsen_lambdasearch_2001}
Thomsen, T.: Lambda-{{Search}} in {{Game Trees}} --- with {{Application}} to {{Go}}. In: Computers and {{Games}}. pp. 19--38. Lecture {{Notes}} in {{Computer Science}}, Springer, Berlin, Heidelberg (2001)

\bibitem{vandenherik_games_2002}
{van den Herik}, H.J., Uiterwijk, J.W.H.M., {van Rijswijck}, J.: Games solved: {{Now}} and in the future. Artificial Intelligence  \textbf{134}(1),  277--311 (Jan 2002)

\bibitem{wang_adversarial_2023}
Wang, T.T., Gleave, A., Tseng, T., Pelrine, K., Belrose, N., Miller, J., Dennis, M.D., Duan, Y., Pogrebniak, V., Levine, S., Russell, S.: Adversarial {{Policies Beat Superhuman Go AIs}}. In: Proceedings of the 40th {{International Conference}} on {{Machine Learning}}. pp. 35655--35739. PMLR (Jul 2023)

\bibitem{wu_relevancezoneoriented_2010}
Wu, I.C., Lin, P.H.: Relevance-{{Zone-Oriented Proof Search}} for {{Connect6}}. IEEE Transactions on Computational Intelligence and AI in Games  \textbf{2}(3),  191--207 (Sep 2010)

\bibitem{wu_game_2024}
Wu, T.R., Guei, H., Wei, T.H., Shih, C.C., Chin, J.T., Wu, I.C.: Game {{Solving}} with {{Online Fine-Tuning}}. In: Advances in {{Neural Information Processing Systems}}. vol.~36 (Feb 2024)

\bibitem{wu_alphazerobased_2021}
Wu, T.R., Shih, C.C., Wei, T.H., Tsai, M.Y., Hsu, W.Y., Wu, I.C.: {{AlphaZero-based Proof Cost Network}} to {{Aid Game Solving}}. In: International {{Conference}} on {{Learning Representations}} (Oct 2021)

\end{thebibliography}
